# MULTILEVEL SENTIMENT ANALYSIS IN ARABIC

1st Ahmed Nassar
*Faculty of Information Technology*
*Islamic university of Gaza*
Gaza, Palestine
aamnassar@iugaza.edu.ps

2nd Ebru Sezer
*Department of Computer Engineering*
*Hacettepe University*
Ankara, Turkey
ebru@hacettepe.edu.tr

*Abstract*—In this study, we aimed to improve the performance results of Arabic sentiment analysis. This can be achieved by investigating the most successful machine learning method and the most useful feature vector to classify sentiments in both term and document levels into two (positive or negative) categories. Moreover, specification of one polarity degree for the term that has more than one is investigated. Also to handle the negations and intensifications, some rules are developed. According to the obtained results, Artificial Neural Network classifier is nominated as the best classifier in both term and document level sentiment analysis (SA) for Arabic Language. Furthermore, the average F-score achieved in the term level SA for both positive and negative testing classes is 0.92. In the document level SA, the average F-score for positive testing classes is 0.94, while for negative classes is 0.93.

## I. INTRODUCTION

The widespread of using social networks, forums, and personal blogs have enabled millions of people to post and share their comments or reviews on the web. Manual collection of huge amount of people's opinions through the use of questionnaires, reviews or comments is time-consuming and it could be impossible. Especially, with the speedy growth of e-commerce. Therefore, the effective solution for this problem is Sentiment Analysis (SA) that is a task of identifying the polarity (positive or negative) of opinions, emotions or evaluations. In general, SA can be defined as it tends to extract the attitude of a person automatically toward some topic or a document [1]. In this study, we bore our interest in sentiment analysis at multi-level such as term level and document level for the Arabic Language. Our objectives include enhancing term level sentiment score by creating a new success formula to compute one prior polarity score for each sentiment term. In other words, Arabic terms has more than one polarity score and we tested different formulas and select the best one to assign one polarity to each term. Secondly, improving the performance results of both term and document levels of Arabic sentiment analysis by investigating the most successfully Machine Learning (ML) classifier to classify the opinion of documents. Finally, implementing the rules with ML approach together to develop obtained SA performance values.

Several factors motivated us to carry out a research in SA in the Arabic Language. The first factor is the consideration of a large number of Arabic audience. The Arabic Language is considered as one of the most widely used top 10 languages on the Internet according to the ranking carried out by the Internet World Stats [2]. Secondly, the Arabic Language has an interesting historical relationship with its people and the region they lived in. Lastly, the large-scale use of the Internet, social media and social networks plays a great and important role on the Arabic Language. There are many studies conducted in SA in the literature but the vast majority of these studies are dedicated to English Language and not directly applicable to Arabic or any other languages as well. Indeed, critical stage of a typical SA is the specification of the polarity. Aaccording to our best knowledge, there are limited studies in Arabic SA that interests the computing of the prior polarity from posterior polarities and we mentioned them in the next section. Herein, some formulas that have not been applied before are proposed to computing prior polarities of lemmas and are re-employed to infer polarity of documents. The authors define clearly feature vectors supplied to ML methods and take into account negations and intensifications. In other words, we describe clear way starting from calculation of terms' polarities to calculate the polarity of document.

ML and Lexicon Based (LB) approaches are two main directions of SA focused by researchers. The ML approach uses linguistic features and applies different classifiers such as Support Vector Machine (SVM), Artificial Neural Network (ANN), and Decision Tree (D-Tree) like as in [3]. The mentioned classifiers also have been used in our study with the consideration of LB approach that depends on known sentiment terms or their roots. Although it is known that building lexicons or dictionaries for Arabic sentiment analysis task is very limited. Rresearchers like Badaro et al. [4], Mahyoub et al. [5] and Mohammad et al. [6] tried to build Arabic sentiment lexicons. Thus, we applied ML approach in our study because of the limitation of LB approach in the Arabic Language.

The rest of this paper is organized as follows: Section II illustrates the previous studies in SA with respect to this field of study. Section III presents the proposed methodology of our study (as shown in Fig.1), then, it discusses the background of ML classifiers with respect to Arabic Language and rules. Section IV illustrates in details the experiments of applying ML approach in both term and document levels with consideration to the application of rules, with a discussion of the study experimental results and evaluations. Conclusions and future work of this study are highlighted in Section V.

## II. RELATED WORK

SA task tends to classify the polarity (positive or negative) in three main levels: the document-level, sentence-level, and the aspect-level [3,7]. However, there is no fundamental difference between sentence and document level classifications because it is possible to say that sentences are just short documents [8]. The SA task differs according to the approach applied, which could be either ML or LB. In the case of applying LB approach like in [9], document sentiment is detected on the basis of adjectives or adverbs phrases by using an unsupervised learning algorithm. Then, semantic orientation is calculated by using Point Mutual Information (PMI). It assigned a label of "recommended" or "not recommended" to the reviews based on the average semantic orientation of the phrases. In the study of [9], the achieved average of accuracy is 74% when the data were sampled from different domains (410 reviews of automobiles, banks, movies, and travel destinations), while the accuracy is 84% for automobile reviews and 66% for movie reviews. Farra et al. [10] used a list of some Arabic words' roots which are extracted by using a stemmer to classify each word in the sentence, by checking against the list of words' roots. If the root is found in the list, its polarity is extracted as a positive, negative, or a neutral. Otherwise, user can add the root of the absent word to the list of learned roots. This study also takes the advantage of both machine learning and lexicon-based approaches by combining them together. As they combined both syntactic and semantic features while classifying the sentiments at Arabic sentence level. They used many features such as frequency of positive, negative and neutral words in each sentence using the semantic list (dictionary) they have built, frequency of contradiction words, frequency of negations such as "ليس، لم، لن، لا" which are negation tools in Arabic and have the same usage of "not" word in English and other features are also used. Guerini et al. [11] used SentiwordNet (SWN) [12] to derive the prior polarity sentiment score from term's posterior polarities thus, they tested many formulas that combine posterior polarities in different ways. They compared the previously used techniques to the proposed ones and incorporated all of them to test whether mixing them can make more improvement to the computing of prior polarity scores. They established motivating basises in computing prior polarity scores. On the other hand, in the case of using ML approach for Arabic Language, Rushdi-Saleh et al. [13] generated new Arabic corpus called OCA, the data of OCA corpus are reviews of movies collected from several movie blogs, obtaining a total of 500 reviews, which entails, 250 positive and 250 negative review response, respectively. Some experiments were also carried out to evaluate the classifiers used in determining the polarity of the review. The best result was achieved by using SVM classifier. They observed that the best result is 90% accurately measured using 10-fold cross validation. El-Halees [14] developed a mechanism for Arabic documents to be classified as positive or negative. His experiments were applied to 1143 posts containing 8793 Arabic statements. The documents were initially entered into a classification model that is based on lexical resources, in which most of the documents were classified at this part of the experiment. After this step, Maximum Entropy (ME) method used the classified documents by the previous model as a training set and then classifies some other documents. Finally, the classified documents from the previous two models were entered to a k-nearest method to classify the remaining documents. On average, his system achieved an accuracy of about 80%, and the F-measure of positive documents has better results than F-measure of negative documents. Shoukry and Rafea [15] investigated ML classifiers (SVM and NB) implementation in Arabic sentence level SA using 1000 tweets from twitter. They used unigram and bigram as features. However, they found that the SVM classifier outperforms the NB classifier in all employed metrics. Moraes et al. [16] made an empirical comparison between SVM and Artificial Neural Networks (ANN) classifiers in the document level of sentiment analysis and they adopted the use of a standard evaluation context with popular supervised methods for feature selection and weighting in a traditional bag of words (BOWs) model. Their work indicated that the results of the experiment of ANN outperform that of SVM for some unbalanced data contexts. Their experimentation included three benchmark data sets (Movie, GPS, Camera and Books Reviews) from amazon.com. In the experiments of movie reviews data set, they indicated that ANN outperformed SVM by a statistically significant difference. They highlighted some limitations that have been rarely discussed in the SA literature which are the computational cost of SVM at the running time and the training time of ANN and they concluded that the reducing of computational effort of both classifiers can be achieved by using Information Gain (a computationally cheap feature selection method) without significantly affecting the resulting classification accuracy. As can be seen, we mostly summarize the studies that are related with Arabic Language and when we compare our approach with these studies, we are contributing them by proposing a general formula to aggregate the posterior polarities of the terms and assigning one polarity for one Arabic term. Additionally, we clearly define employed and proposed vectors while usage of ML methods. Finally, we propose the clear way that is interpretable and repeatable to find out sentiments of documents in Arabic Language.

III. BACKGROUND AND ML CLASSIFIERS

ML approach is one of the main directions of SA focused by researchers as mentioned before, which applies different classifiers and uses linguistic features [3]. In sentiment classification of text, employed vectors hold on some features that are collected from a document or a sentence and ML methods uses these vectors to specify the polarity and sentiment. Actually, we keep going this approach in general and we proposed the most useful vector definitions for this type of application. In this section, employed ML methods in this study are introduced (as shown in Fig.1), and an overview of Arabic Language is presented. Finally, the rules that entail the negation and intensification of the Arabic Language are discussed.

*A. Support Vector Machine Classifier (SVM)*

Support Vector Machine is a supervised learning algorithm with many popular specialties that motivated many researchers to use it. Moreover, it is widely applied in text classification and SA problems due to its superiority over other classifiers [16]. The main idea of SVM is based on the best separation of different classes by using determined linear separator. According to Joachims [17], text data can be classified by using SVM because of the sparse nature of the text. However, some of the text features can be irrelevant but they tend to be organized into categories that can be linearly separated. Due to this fact, SVM have been frequently employed in SA field. For example Li and Li [18], and we also used SVM as a sentiment polarity classifier to make our results comparable with the previously results. In other words, we prefer to demonstrate the usefulness of our formulas and feature vectors by using common classification methods of SA domain. In this study, LIBSVM software package [19] has been used with its default parameters values as implemented in the Matlab software. Moreover, the usual nonlinear kernel Redial Basis Function (RBF) which is used by Moraes et al. [16], is also used to train all the SVM models, as it has better performance than other kernel functions performance in the experiments of our study.

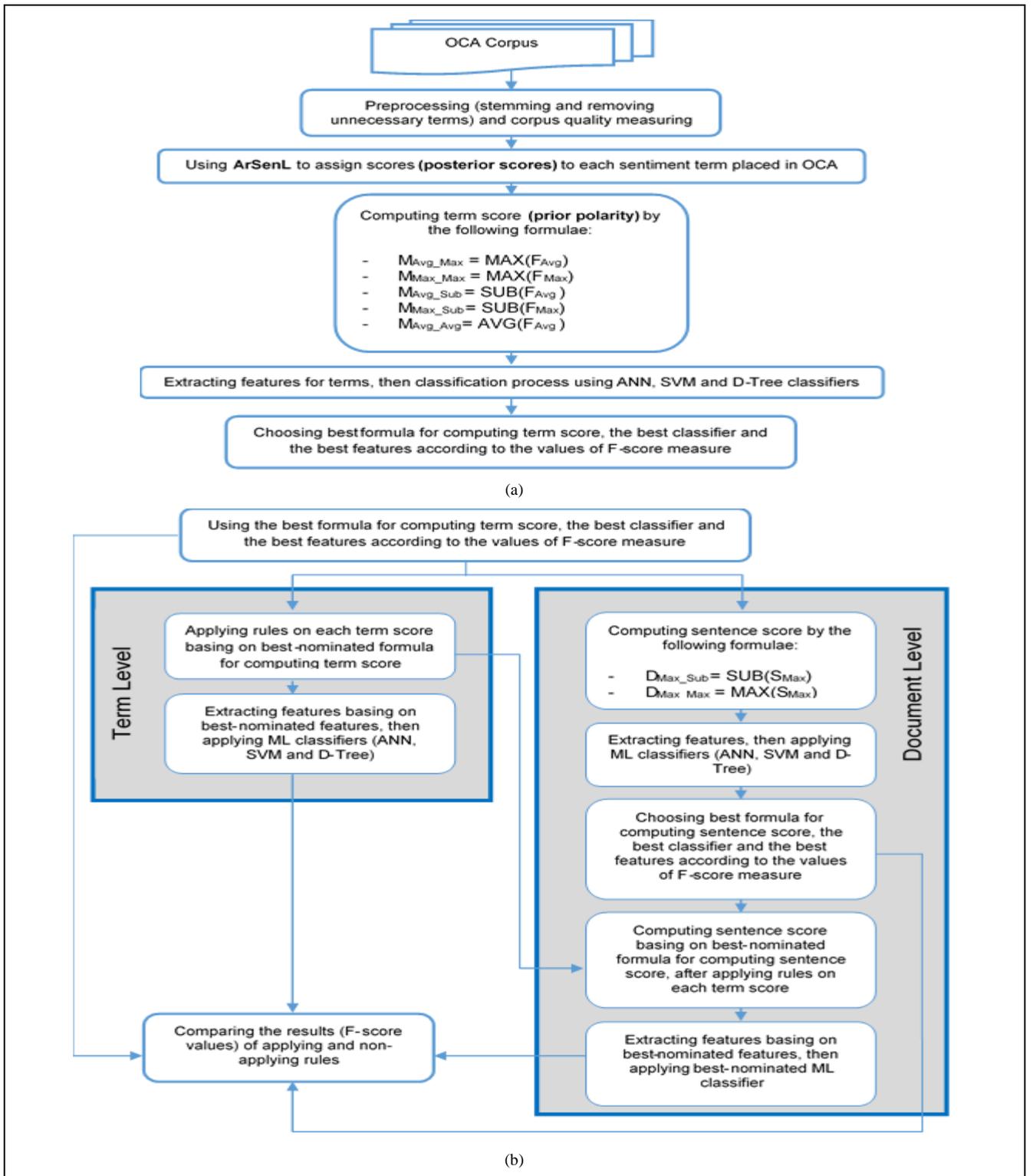

Fig. 1 Proposed methodology of multi-level SA in Arabic Language using ML approach : (a) Preprocessing and nominating the best formula for computing term score , (b) Using the nominated formula for both term and document levels

## B. Artificial Neural Network Classifier (ANN)

Artificial Neural Network includes a collection of artificial neurons in which the neuron is the basic unit of ANN. There are input and output for ANN, the inputs are represented by the vector $\overline{A_i}$. Each neuron has a set of weights that are represented by W. The ANN can be represented by the linear function: $P_i = W * \overline{A_i}$. In binary classification, the sign of predicted function (Pi) yields Bi, which represents the class label of $\overline{A_i}$. According to Medhat et al. [3], multilayer ANN is more complex and the training process is difficult. This is due to back propagation process, at which errors back-propagated throw different layers. However, multilayer ANN is still used for nonlinear boundaries. For the ANN classifier applied in this study, traditional feed-forward network with single hidden layer that includes 15 neurons was applied. The training process for each model is repeated more than three times to avoid the problem of convergence to a satisfactory solution [20]. Additionally, gradient descent with momentum and adaptive learning rate back-propagation (traingdx) has been used as implemented in the Matlab software.

*C. Decision Tree Classifier (D-Tree)*

D-Tree classifier has a hierarchical structure that includes attributes or nodes which represent training data space, each condition on the attribute value is used to divide the nodes or the data [21]. In text classification field, the condition can be about the absence or the presence of the word. The division of nodes is continued in recursive fashion until certain minimum numbers of records contained in leaf nodes are found. These records can be used for the classification purpose. However, there are D-Tree implementations for text classification task, such as J48 implementation in the Weka Data Mining tool, which is based on the C4.5 D-tree algorithm. Jia et al. [22] built a system at the document level in which the document is classified to be opinionated in the case of finding one sentiment sentence in it. Unigrams and bigrams features were selected using Chi-square test at the sentence level. Then a D-Tree classifier was used to determine the polarity of opinionated documents. In this study, the default parameters values of D-tree were used as implemented in the Matlab software. For example, the confidence factor value is 0.25 and the value of the minimum number of instances per leaf is 2.

*D. Arabic Language Overview*

The Arabic Language is considered as one of the top 10 languages mostly used on the Internet [2]. It is spoken by millions of people, making it one of the five most spoken languages in the World, and also one of the formal languages of United Nations (UN) organization [23]. The Arabic Language was spoken in Hejaz and its surrounding areas before the Islamic age. In addition, it is the language of Quran (The Holy Islamic Book), so it became the official language in the Islamic regions. Many Muslims from different cultures learned this language in order to understand Islamic religion deeply. The orientation of writing is from right to left. The Arabic alphabet consists of 28 letters, and can be extended to 90 elements by writing additional shapes, marks, and vowels [24]. There are two forms of Arabic Language; (i) The Modern Standard Arabic (MSA) and the (ii) Dialectal Arabic. In Arabic Countries, Modern Standard Arabic (MSA) is derived from the language of the Quran. In addition, it is commonly used in books, newspapers, media, formal speeches, movie reviews, and so on. MSA largely follows the grammatical standards of Quranic Arabic, it uses much of the same vocabulary. However, it has discarded some grammatical constructions and vocabulary that no longer have any counterpart in the spoken varieties. Additionally, it has adopted certain new constructions and vocabulary from the spoken varieties [23]. While Dialectal Arabic includes all forms of nowadays currently spoken Arabic. It deviates from the Standard Arabic to some extent [13].

The Arabic Language is one of the Semitic Languages in which the morphology is complex and unusual, that is, it employs the approach of constructing words from a basic root. The Arabic Language has a non-concatenative "root-and-pattern" morphology; A root consists of a set of bare consonants (usually three). In Arabic Language, different patterns can be used to form numerous words. For example, from a single root **k-t-b**, numerous words can be formed such as; **K**at**ab**tu: 'I wrote'; A**kt**ab**tu: 'I dictated'; A**kt**ub**u: 'I write'; **Kutib**a: 'it was written', and so on. This research is concerned with MSA. The challenge is that not all natural language processing approaches that are applicable to most languages can be applied to the Arabic Language. The text needs much preprocessing before some of the methods can be applied.

*E. Rules*

Rules designed for Arabic Language include two main situations that may change the meaning of a word or a sentence; these are "intensification" and "negation" tools. For example, the intensification tools in Arabic including the words "إفراط، جدا، كبيرا، مطلق" can have the same usage of the word "very" in the English language. On the other hand, the negation tools "ليس، لم، لن، لا" in Arabic have the same usage of the word "not" in the English language. In our study, it is decided that for each term: (i) in case of positive terms, if the term has intensification word before or after it, its score will be increased to 1, while in case of negative terms, the term score is decreased to -1. (ii) If the term has negation word before it, its score will be negated, in which the positive term becomes a negative term and the negative term becomes a positive term.

## IV. SA MODELS AND EXPERIMENTS

In this section, our experiments include application and non-application of rules with ML approach in order to identify their effect on results. This section is ordered as follows: Section A presents preprocessing task of OCA corpus including (i) measuring the corpus quality by comparing it with the other Arabic corpora using the same quality measures. (ii) stemming and identifying the sentiment scores of each term placed in OCA (iii) computing one sentiment score (prior polarity) for each term and (iv) computing sentiment score of sentences by using proposed formulas. Section B illustrates ML based classifications with the details of extracted features for term and document levels and the effect of rules on the classification. Section C summarizes the performance of evaluators used to assess experimental results.

*A. Preprocessing of the Data Set*

OCA corpus is a new Arabic resource that is made available to the Scientific Community to be used in sentiment analysis task . Our experiments were conducted on OCA data set similarly with the study of [13]. Two samples of positive and negative reviews of the OCA corpus are given in Table I to be more clear. Preprocessing for OCA dataset is made by removing useless terms (i.e. spaces, punctuation marks, numbers, etc.) in SA task. Table II presents statistical definition of the OCA corpus after preprocessing.

TABLE I. Samples of positive and negative OCA reviews

| | |
|---|---|
| Positive review | نعم إنه فيلم "الحسناوات" ولكن موضوعه ذكي ، كما أنه مؤثر وسيثير لديك رغبة شديدة في البكاء ، إلى جانب أنه مرح وعلى درجة عالية من الإخراج والتمثيل.<br>إنه يستحق ثمن التذكرة والساعتين اللتين ستضيعهما من وقتك الثمين على مشاهدته.<br>التقييم العام : 3<br>In Her Shoes Yes, it is the film "Belles" But the theme of intelligent, influential as it is, and will raise you have a strong desire to cry, to the side because it's funny and a high degree of output and representation. It is worth the price of the ticket and two hours of your precious time to watch it. User Rating: 3 |
| Negative review | اسم الفيلم: 7 ورقات كوتشينة<br>إسم الكاتب: د.صلاح الغريب<br>من وجهة نظري الشخصية الفيلم يعتبر ضعيف جداً جداً من جميع النواحى ...<br>ولم يلفت نظري سوى الأداء الجيد للفنان ( محمد سليمان ) و الفنانة ( رانيا شاهين ) .<br>التقييم: 1/10<br>Movie Name: 7 papers Kuchinp Author Name: Dr. Salah strange from the point of my personal film is very, very weak in all respects ... Did not draw my eyes only good performance of the artist (Mohammad Suleiman) and the artist (Rania Shaheen). Rating: 1 / 10 |

TABLE II. Statistical details of OCA after pre-processing

| | Positive | Negative |
|---|---|---|
| **Total No. of documents** | 250 | 250 |
| **Total No. of tokens (words)** | 77704 | 60748 |
| **Total No. of unique tokens in OCA** | 5984 | 5812 |

*1) Quality Measurement of OCA Corpus*

First of all, we aimed to measure OCA corpus quality before its usage in the task of sentiment analysis. Thus, we used Zipf's law to measure the quality of OCA corpus as it is presented in the previous studies by Benajiba and Rosso [25] and Yahya and Salhi [26]. Zipf's law states that: "in a corpus, the frequency of a word is inversely proportional to its rank"; Zipf's law can be expressed by Eq.1

$$F = C/r^a \qquad (1)$$

Where F is the frequency of words, a is a constant close to 1, r is the rank of the word and C is the highest observed frequency. According to this approach, the most frequent word occurs twice as often as the second most frequent word, and the third most frequent word occurs one-third of the most frequent word and so on. In our experiments, a relationship has been drawn between actual frequency and the rank using the logarithmic measure as presented in Fig. 2a. The relationship in Fig. 2b, is also drawn between ideal (Zipf's) frequency and the rank using logarithmic measure. When a comparison was made between actual and ideal frequencies as seen in Fig. 3, it is concluded that few words are often used and many of the words are rarely used.

To assess the adherence of word frequencies (i.e. the word frequency distribution of text to Zipf's law distribution), we used Kullback-Liebler distance measure ($D_{KL}$), this distance is asymmetric and measures the distance from a "true" probability distribution P (Zipf's law distribution based on rank) to an arbitrary probability distribution Q (word frequency distribution) as discussed by Benajiba and Rosso [25]. Eq.2 can express this function:

$$D_{KL}(P, Q) = \sum_i P(i) . \log P(i) / Q(i) \qquad (2)$$

Where P is Zipf's law distribution and Q is word frequency distribution. For OCA corpus the $D_{KL}$ obtained has been compared to the $D_{KL}$ resulted from [25] study as presented in Table III. Furthermore, it is also compared to [26] study as presented in Table IV.

TABLE III. Comparison between $D_{KL}$ for OCA and $D_{KL}$ for Arabic corpora by Rosso and Benajiba [18]

| Corpus | Kullback-Liebler Distance |
|---|---|
| **OCA** | 18980.41 |
| **Abu-Taïb AlMoutanabbi poetry** | 22120.32 |
| **Newspaper articles** | 32836.98 |
| **A Linux Red Hat installation tutorial book** | 41983.44 |
| **Religious book of the Imam Ibnu Qayyim El Jawziyah** | 28870.38 |

TABLE IV. Comparison between $D_{KL}$ for OCA and $D_{KL}$ for Arabic corpora by Salhi and Yahya [19]

| Corpus | Kullback-Liebler Distance |
|---|---|
| **OCA** | 0.037424 |
| **Al-Quds** | 0.084 |
| **Ar. Wikipedia** | 0.100 |
| **Computing** | 0.143 |
| **Economics** | 0.134 |
| **History** | 0.134 |
| **Literatures** | 0.132 |
| **Physics Related** | 0.139 |
| **Medicine Related** | 0.136 |
| **Politics** | 0.128 |
| **Religions** | 0.139 |
| **Sports** | 0.106 |

According to the comparisons between our study result and the results of previous studies regarding $D_{KL}$, we concluded that the OCA corpus has the best $D_{KL}$. Thus, OCA data set quality outperforms other Arabic data sets quality. For this reason, we can dependably use it in SA task.

*2) Stemming Process and Identifying Sentiment Terms*

After the preprocessing phase of OCA corpus, the opinion terms from the corpus are decided through stemming all OCA terms. This is done by using the commonly used Standard Arabic Morphological Analyzer (SAMA) [27]. This analyzer can produce all possible reading out of context for a given word. Whereas each word lemma has English gloss and part of speech tag. After stemming process, sentiment scores for each term root placed in OCA is given by comparing them to large-scale Arabic sentiment analysis lexicon (ArSenL). ArSenL is a large sentiment lexicon for the Arabic Language in which each word seed (lemma) has more than one score (or posterior polarities). The study of [4] created ArSenL by using Arabic and English resources: (i) English SentiWordNet (ESWN) [12]; (ii) English WordNet (EWN) [28]; (iii) Arabic WordNet (AWN) [29] and (iv) Standard Arabic Morphological Analyzer (SAMA) [27].

The English synsets were connected to the lemma entries in the Arabic resources. Noting that, each word root (lemma) in the lexicon have many senses or scores which are rated in the interval of [-1..1]. For example, Table V presents total five ArSenL positive and negative scores for the word: "hot" (ساخن). In this study, we used ArSenL to derive one numerical score from positive scores and one numerical score from negative scores. Then, we formulated aggregation formulae

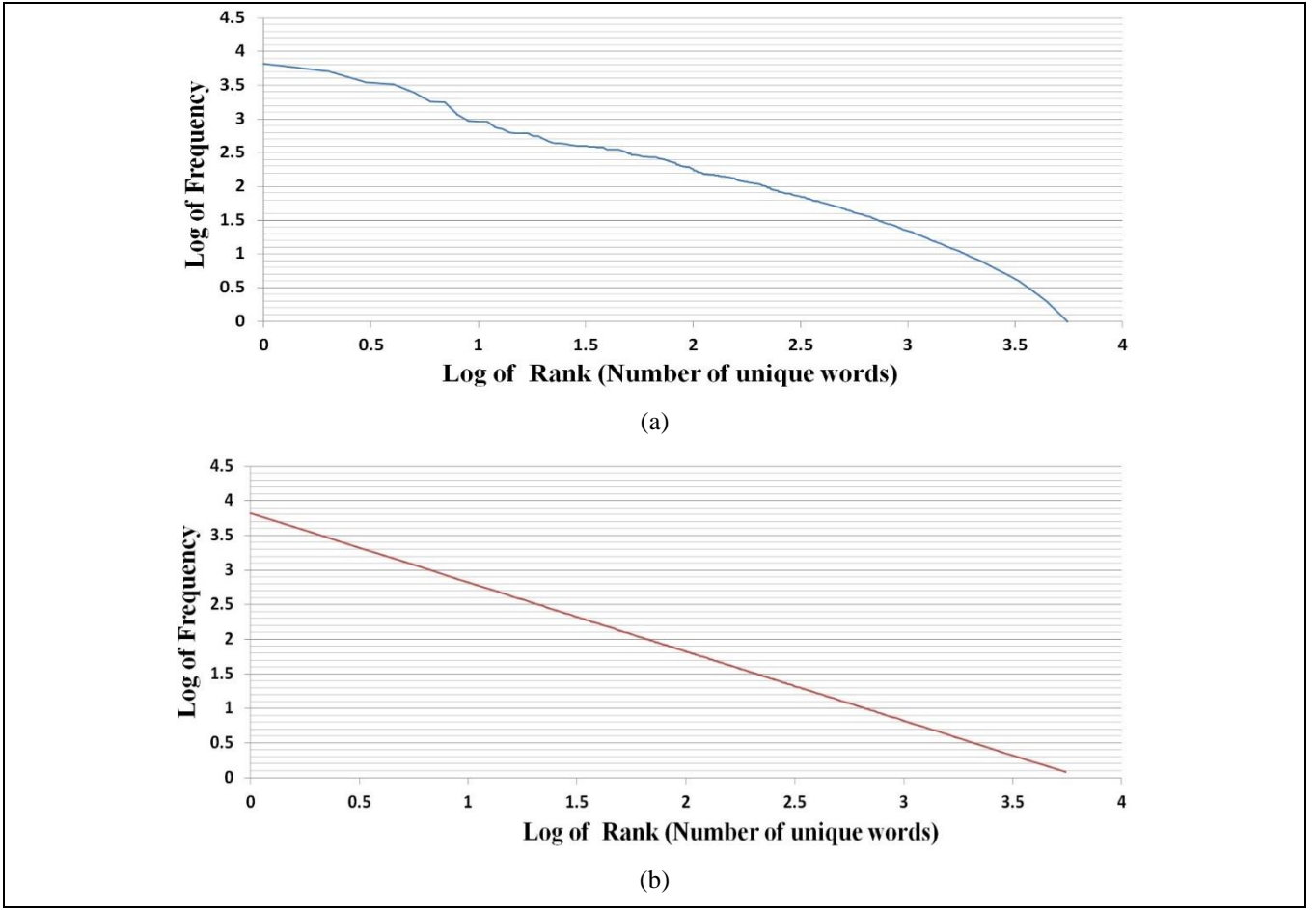

Fig. 2 (a) OCA frequencies VS Rank, (b) Ideal Frequency VS Rank

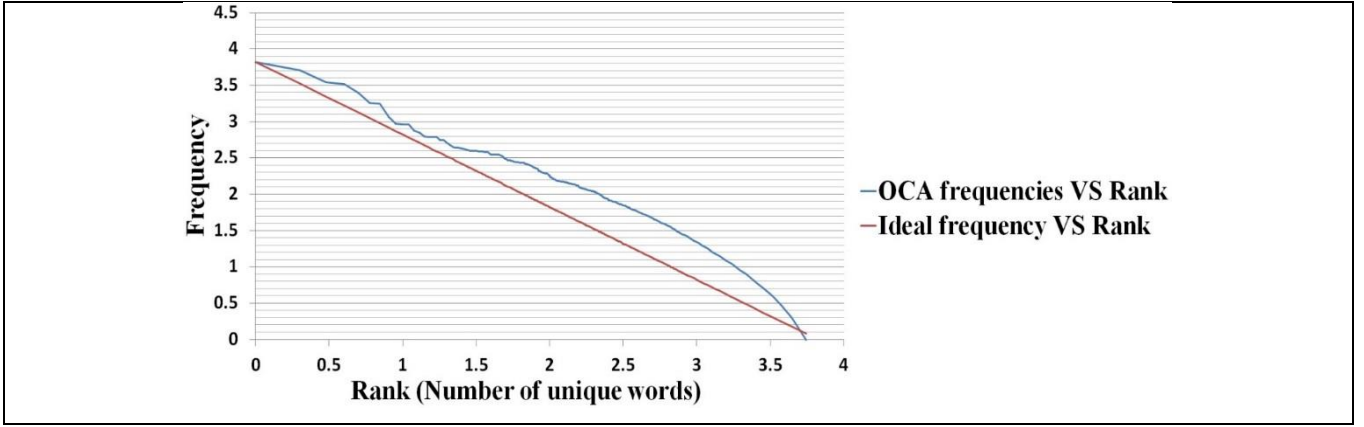

Fig. 3 Comparison between Actual and Ideal frequencies VS Rank

to derive one score (or prior polarity) from the posterior scores for each term placed in OCA as illustrated in the next section.

TABLE V. Five ArSenL positive and negative scores for the term: "hot" (ساخن)

| Lemma by SAMA | Positive scores | Negative scores |
|---|---|---|
| sAxin | 0.375 | 0.25 |
| sAxin | 0.75 | 0.125 |
| sAxin | 0.5 | 0.375 |
| sAxin | 0.25 | 0.25 |
| sAxin | 0.125 | 0 |

*3) Prior Polarity Computing Formulae for Term Level*
In this phase, one prior sentiment score is extracted from posterior positive or negative scores as shown in Fig 1. However, total 5 formulae (or equations) are constructed to derive one prior sentiment score from posterior scores in two steps. In the first step, two scores which are F (Positive Score) and F (Negative Score) are derived from posterior scores for each word lemma using Eq.3 and Eq.4 in Table VI. Secondly, we derived unique prior polarity score from the resulted two scores of $F_{Avg}$ and $F_{Max}$ separately by using the Equations 5 – 9 in Table VI.

TABLE VI. Prior polarity computing equations

| The equation | Discerption | |
|---|---|---|
| $F_{Avg}$ = (Avg (\|Positive Score\|), Avg (\|Negative Score\|)) | $F_{Avg}$ takes the average of the absolute value of positive and negative scores independently for each term. | (3) |
| $F_{Max}$ = (Max (\|Positive Score\|), Max (\|Negative Score\|)) | $F_{Max}$ takes the maximum of the absolute value of positive and negative scores independently for each term. | (4) |
| $M_{Avg\_Max}$ = MAX($F_{Avg}$) | $M_{Avg\_Max}$ takes the maximum score of the resultant two scores from $F_{Avg}$. | (5) |
| $M_{Max\_Max}$ = MAX($F_{Max}$) | $M_{Max\_Max}$ takes the maximum score of the resultant two scores from $F_{Max}$. | (6) |
| $M_{Avg\_Sub}$ = SUB($F_{Avg}$) | $M_{Avg\_Sub}$ takes the difference of the resultant two scores from $F_{Avg}$. That is, the subtraction of the negative score from the positive score. | (7) |
| $M_{Max\_Sub}$ = SUB($F_{Max}$) | $M_{Max\_Sub}$ takes the difference of the resultant two scores from $F_{Max}$. That is the subtraction of the negative score from the positive score. | (8) |
| $M_{Avg\_Avg}$ = AVG($F_{Avg}$) | $M_{Avg\_Avg}$ takes the average of the resultant two scores from $F_{Avg}$. | (9) |

In Eq.9, the negative sign is added to the negative score before taking the average. In addition, in Eq.5 and Eq.6, it is important to note that, a negative sign is added to the final unique prior score if it was negative before taking the absolute value. To nominate the best of the five equations (or formulae) of deriving prior polarity, five OCA files based on the five formulae ($M_{Avg\_Max}$, $M_{Max\_Max}$, $M_{Avg\_Sub}$, $M_{Max\_Sub}$ and $M_{Avg\_Avg}$) were created. After that, we tested each OCA file independently using ML methods such as ANN, D-Tree and SVM classifiers. Furthermore, the $M_{Max\_Sub}$ formula has the best results amongst all other formulae, in which the average F-score achieved in the term level for both positive and negative testing classes is 0.92. Thus, the $M_{Max\_Sub}$ formula is certified for giving one score for each term placed in OCA corpus.

*4) Sentence Score Computing Formulae*

One score for every sentence in each of the OCA documents has been computed. $M_{Max\_Sub}$ formula has been certified for giving one score for each term of the OCA corpus. However, two versions of the OCA corpus based on the sentence score have been made. Whereas, in the first version each sentence score is computed applying rules, and the second version was computed without applying rules. In order to make comparisons and to identify the effectiveness of the rules on the experiments result, we firstly derived two scores S (Positive term score) and S (Negative term score) using Eq.10 in Table VII. Secondly, one sentence score is derived from the result of two scores of $S_{Max}$ formula by using Eq.11 and Eq.12 in Table VII. To nominate the best of the two equations (Eq.11 and eq.12) of deriving one sentence score, we tested each OCA file independently using ML methods and, the $D_{Max\_Max}$ formula has results better than the results of $D_{Max\_Sub}$ formulae.

TABLE VII. Sentence score computing equations

| The equation | Discerption | |
|---|---|---|
| $S_{Max}$ = (Max (\|Positive term Score\|), Max (\|Negative term Score\|)) | $S_{Max}$ takes the maximum absolute value of the positive term score in the sentence and the maximum absolute value of the negative term score in the sentence. | (10) |
| $D_{Max\_Sub}$ = SUB($S_{Max}$) | $D_{Max\_Sub}$ takes the difference of the resultant two scores from $S_{Max}$. That is the subtraction of the negative score from the positive score. | (11) |
| $D_{Max\_Max}$ = MAX($S_{Max}$) | $D_{Max\_Max}$ takes the maximum score of the resultant two scores from $S_{Max}$, and the negative sign is added to the final sentence score in case of the negative term is the maximum. | (12) |

*B. Arabic Sentiment Analysis Using Machine Learning Approach*

According to the study methodology that is presented in Fig.1 previously, features were selected for both term and document levels independently, considering the application and non-application of rules on both levels. Afterward, we applied ML approach on the term and the document level of the OCA corpus. Finally, we used three encouraging ML classifiers from SA literature, which are SVM, ANN and D-Tree to classify OCA documents.

*1) Features Extraction*

Based on the $M_{Max\_Sub}$ formula for computing prior polarity for each sentiment term placed in OCA, features were extracted in two levels of SA, the term, and the document level. Where, we made two versions of features for each term level and document level. After which, the first version includes all features that were created with the application of intensification and negations tools (rules) on each term placed in OCA. While in the second version of features, the rules were not applied on each term placed in OCA.

*a) Term Level Features*

In accordance with the previous sub section 3, five OCA files were built, in each file every word is assigned a prior score according to the $M_{Max\_Sub}$ formula. Each document in the file is represented by a vector of total eight features. These features are ordered as: count of positive scores, count of negative scores, summation of positive scores, summation of negative scores, average of positive scores, average of negative scores, first subjective score and last subjective score. For example, Table VIII gives the sample for one of the OCA files including the total eight features. In order to ensure the effectiveness of our features and the testing methods, we omitted the last two feature, that is, the first subjective score and last subjective score, thus, we have total six features. Then, the same experiments were repeated for the rest of the scores.

*b) Document Level Features*

With respect to the previous sub section 4, two OCA files were built, each file sentence is assigned one score according to the $D_{Max\_Max}$ and $D_{Max\_Sub}$ formulae. Total seven features were selected for each OCA file independently which were ordered as: count of positive sentences, count of negative sentences, maximum score of positive sentences, maximum score of negative sentences, first sentence score, middle sentence score and last sentence score. Table IX presents the sample for two OCA files including the total seven features. In order to ensure the effectiveness of the features and the testing methods, we, first of all, omitted the third and the fourth features, which are the maximum score of positive sentences and the maximum score of negative sentences, thus we have total five features. Then, the same experiments were repeated for the rest of the scores. Furthermore, we omitted the last three features, which are the first sentence score, middle sentence score, and last sentence score, thus we have total four features. Then, the same experiments were also repeated for the rest of the scores.

*2) Classification Task*

In the classification task, we put the features vector into a format that can be processable by each of SVM, ANN and D-Tree classifiers. According to this study experiments, each classifier settings are presented in the previous subsections (i.e. A, B and C) of Section III. Some classifiers need parameter settings which are still a research issue [16]. Moreover, we applied 5-fold cross validation for testing and validation purpose to ensure that every sample of OCA corpus has the same chance of appearing in training and validation set [30]. Indeed, the preprocessed OCA corpus is divided into 5 testing files in which each file includes 100 documents in which 50 documents are positive and 50 are negative. The rest corpus is divided into 5 training files, so that each file contains 400 documents (200 positive and 200 negative documents).

*C. Evaluation Measures*

The objective of classification task is to find the best approach and formula for Arabic sentiment analysis task, and to show how classification results are improved when the number of features is increased with consideration of application and non-application of the rules.

We used three commonly used metrics in the literature by Badaro et al. [4], El-Halees [14] and Al-Azani and El-Alfy [31] which are Precession, Recall, and F-score. The equations representing these metrics are listed in Table X. The average of F-scores is used in comparing all the results of the classification in our experiments.

TABLE VIII. Sample of one of the OCA files including the total eight features in term level

| Class | 0 | 0 | 0 | 0 | 1 | 1 | 1 |
|---|---|---|---|---|---|---|---|
| **CountPos** | 166 | 197 | 110 | 19 | 61 | 37 | 45 |
| **CountNeg** | 141 | 147 | 66 | 21 | 21 | 32 | 27 |
| **SumPos** | 20.2 | 25.51 | 16.24 | 2.61 | 5.56 | 5.16 | 8.29 |
| **SumNeg** | -18.8 | -21.9 | -11.9 | -4.04 | -3.7 | -6 | -3.7 |
| **AvgPos** | 0.12 | 0.13 | 0.15 | 0.14 | 0.09 | 0.14 | 0.18 |
| **AvgNeg** | -0.13 | -0.15 | -0.18 | -0.19 | -0.2 | -0.2 | -0.1 |
| **F_Subj** | 0.04 | 0.04 | -0.08 | -0.56 | 0.04 | 0.04 | 0.04 |
| **L_Subj** | 0.21 | -0.25 | 0.03 | 0.04 | 0.02 | 0.25 | 0.26 |

TABLE IX. Sample of one of the OCA files including the total seven features at the sentence level

| Class | 0 | 0 | 0 | 1 | 1 | 1 | 1 |
|---|---|---|---|---|---|---|---|
| **CountPos** | 8 | 38 | 6 | 1 | 1 | 1 | 1 |
| **CountNeg** | 2 | 22 | 1 | 0 | 0 | 0 | 0 |
| **MaxPos** | 0.75 | 1 | 0.88 | 0.75 | 0.75 | 0.88 | 0.75 |
| **MaxNeg** | -0.75 | -0.88 | -0.5 | 0 | 0 | 0 | 0 |
| **FirstScore** | -0.25 | 0.13 | 0.5 | 0.75 | 0.75 | 0.88 | 0.75 |
| **MiddleScore** | -0.75 | 0 | 0.75 | 0.75 | 0.75 | 0.88 | 0.75 |
| **LastScore** | 0.13 | 0.13 | 0.13 | 0.75 | 0.75 | 0.88 | 0.75 |

TABLE X. Evaluation metrics

| Evaluation measures | The equations | |
|---|---|---|
| Precession for positive classes | $\text{Precession} = \frac{\text{true\_positive}}{\text{true\_positive} + \text{false\_positive}}$ | (13) |
| Precession for negative classes | $\text{Precession} = \frac{\text{true\_negative}}{\text{true\_negative} + \text{false\_negative}}$ | (14) |
| Recall for positive classes | $\text{Recall} = \frac{\text{true\_positive}}{\text{true\_positive} + \text{false\_negative}}$ | (15) |
| Recall for negative classes | $\text{Recall} = \frac{\text{true\_negative}}{\text{true\_negative} + \text{false\_positive}}$ | (16) |
| F-score | $\text{F-score} = \frac{2 * \text{Precision} * \text{Recall}}{\text{Precision} + \text{Recall}}$ | (17) |

V. RESULTS

In this section, we present the results of applying ML approach on both term and document levels of OCA corpus, with application and non-application of rules. The first obtained results are Term Level SA without rules and all of them are listed in Table XI-XIII. By comparing the results of the three ML classifiers in the tables, three conclusions were reached. Firstly using 8-features performs better than using 6-features in the three tables. The second conclusion is that the results of ANN classifier in Table XI outperforms both the results of SVM classifier in Table XII and that of D-Tree classifier in Table XIII. Lastly, $M_{Max\_Sub}$ formula has the best results amongst the employed formulae in Table XI in which the average F-score achieved in the term level for positive testing classes was **0.92**, and same value was obtained for the negative classes as well. So the $M_{Max\_Sub}$ formula is credited to be advised in order to get better one score for each Arabic sentiment term.

Results of term Level SA employing rules are presented in Table XIV. Where the results of the ANN classifier outperform the results of the SVM and D-Tree classifiers. The results of non-application of rules in Table XI outperform the results of application of rules in Table XIV specifically in the case of applying $M_{Max\_Sub}$ formula and a total of 8 features in both experiments. For example, the average F-score in positive and negative classes for testing data in Table XI is **0.92.** While the average F-score in positive and negative classes for testing data in Table XIV is **0.91.**

TABLE XI. Average of performance by using total 8 and 6 features with ANN classifier

| | | Positive classes | | Negative classes | |
|---|---|---|---|---|---|
| | Formulae | F-score for training data | F-score for testing data | F-score for training data | F-score for testing data |
| **8 Features** | $M_{Max\_Max}$ | 0.78 | 0.89 | 0.76 | 0.88 |
| | $M_{Avg\_Max}$ | 0.79 | 0.87 | 0.76 | 0.87 |
| | $M_{Avg\_Sub}$ | 0.77 | 0.84 | 0.73 | 0.84 |
| | $M_{Max\_Sub}$ | 0.81 | **0.92** | 0.79 | **0.92** |
| | $M_{Avg\_Avg}$ | 0.77 | 0.86 | 0.75 | 0.85 |
| **6 Features** | $M_{Max\_Max}$ | 0.73 | 0.80 | 0.72 | 0.79 |
| | $M_{Avg\_Max}$ | 0.75 | 0.84 | 0.71 | 0.81 |
| | $M_{Avg\_Sub}$ | 0.72 | 0.80 | 0.70 | 0.80 |
| | $M_{Max\_Sub}$ | 0.74 | 0.78 | 0.71 | 0.77 |
| | $M_{Avg\_Avg}$ | 0.74 | 0.82 | 0.70 | 0.81 |

TABLE XII. Average of performance by using total 8 and 6 features with SVM classifier

| | Formulae | Positive classes | | Negative classes | |
|---|---|---|---|---|---|
| | | F-score for training data | F-score for testing data | F-score for training data | F-score for testing data |
| 8 Features | $M_{Max\_Max}$ | 0.70 | 0.58 | 0.48 | 0.48 |
| | $M_{Avg\_Max}$ | 0.68 | 0.61 | 0.47 | 0.49 |
| | $M_{Avg\_Sub}$ | 0.67 | 0.63 | 0.53 | 0.39 |
| | $M_{Max\_Sub}$ | 0.74 | 0.60 | 0.60 | 0.56 |
| | $M_{Avg\_Avg}$ | 0.66 | 0.68 | 0.51 | 0.53 |
| 6 Features | $M_{Max\_Max}$ | 0.67 | 0.55 | 0.45 | 0.47 |
| | $M_{Avg\_Max}$ | 0.67 | 0.60 | 0.45 | 0.43 |
| | $M_{Avg\_Sub}$ | 0.67 | 0.63 | 0.51 | 0.41 |
| | $M_{Max\_Sub}$ | 0.74 | 0.57 | 0.57 | 0.55 |
| | $M_{Avg\_Avg}$ | 0.66 | 0.66 | 0.49 | 0.57 |

TABLE XIII. Average of performance by using total 8 and 6 features with D-Tree classifier

| | Formulae | Positive classes | | Negative classes | |
|---|---|---|---|---|---|
| | | F-score for training data | F-score for testing data | F-score for training data | F-score for testing data |
| 8 Features | $M_{Max\_Max}$ | 0.75 | 0.75 | 0.69 | 0.76 |
| | $M_{Avg\_Max}$ | 0.76 | 0.77 | 0.72 | 0.76 |
| | $M_{Avg\_Sub}$ | 0.72 | 0.75 | 0.69 | 0.77 |
| | $M_{Max\_Sub}$ | 0.72 | 0.77 | 0.72 | 0.78 |
| | $M_{Avg\_Avg}$ | 0.75 | 0.76 | 0.68 | 0.77 |
| 6 Features | $M_{Max\_Max}$ | 0.74 | 0.67 | 0.58 | 0.61 |
| | $M_{Avg\_Max}$ | 0.74 | 0.69 | 0.58 | 0.62 |
| | $M_{Avg\_Sub}$ | 0.69 | 0.68 | 0.56 | 0.60 |
| | $M_{Max\_Sub}$ | 0.71 | 0.70 | 0.58 | 0.60 |
| | $M_{Avg\_Avg}$ | 0.73 | 0.72 | 0.56 | 0.61 |

Results of document Level SA without rules are presented in Table XV and Table XVI. By comparing the results of the three ML classifiers on these tables, we made three conclusive statements. The first conclusion is that using total 7 features is better than using a total of 5 or 4 features in both tables. The second conclusion is that ANN gives better results than SVM and D-Tree in both tables, which was based on a total of 7 features. The last conclusive statement is that the ANN classifier results based on the $D_{Max\_Max}$ formula in Table XV have better result than the ANN classifier's results based on $D_{Max\_Sub}$ formula in Table XVI. Therefore, the $D_{Max\_Max}$ formula is recommended to be used when the need to getting one score for each Arabic sentiment sentence is needed.

TABLE XIV. Average of performance for using total 8 features in case of applying rules with the ML classifiers in the term level based on $M_{Max\_Sub}$ formula

| Classifier | Positive classes | | Negative classes | |
|---|---|---|---|---|
| | F-score for training data | F-score for testing data | F-score for training data | F-score for testing data |
| **ANN** | 0.78 | **0.91** | 0.77 | **0.91** |
| D-Tree | 0.71 | 0.80 | 0.65 | 0.80 |
| SVM | 0.66 | 0.58 | 0.54 | 0.37 |

TABLE XV. Average of performance of using total 7, 5 and 4 features based on $D_{Max\_Max}$ formulae in the document level without rules

| | Classifier | Positive classes | | Negative classes | |
|---|---|---|---|---|---|
| | | F-score for training data | F-score for testing data | F-score for training data | F-score for testing data |
| 7 Features | **ANN** | 0.78 | **0.93** | 0.78 | **0.92** |
| | D-Tree | 0.71 | 0.87 | 0.68 | 0.88 |
| | SVM | 0.60 | 0.79 | 0.63 | 0.82 |
| 5 Features | ANN | 0.77 | 0.93 | 0.77 | 0.92 |
| | D-Tree | 0.76 | 0.87 | 0.71 | 0.88 |
| | SVM | 0.61 | 0.82 | 0.63 | 0.85 |
| 4 Features | ANN | 0.67 | 0.87 | 0.60 | 0.84 |
| | D-Tree | 0.65 | 0.75 | 0.58 | 0.76 |
| | SVM | 0.58 | 0.79 | 0.57 | 0.83 |

TABLE XVI. Average of performance of using total 7, 5 and 4 features based on $D_{Max\_Sub}$ formulae in the document level without rules

| | Classifier | Positive classes | | Negative classes | |
|---|---|---|---|---|---|
| | | F-score for training data | F-score for testing data | F-score for training data | F-score for testing data |
| 7 Features | **ANN** | 0.73 | **0.90** | 0.70 | **0.89** |
| | D-Tree | 0.64 | 0.81 | 0.64 | 0.88 |
| | SVM | 0.59 | 0.79 | 0.62 | 0.82 |
| 5 Features | ANN | 0.71 | 0.86 | 0.71 | 0.87 |
| | D-Tree | 0.64 | 0.80 | 0.65 | 0.86 |
| | SVM | 0.60 | 0.78 | 0.60 | 0.81 |
| 4 Features | ANN | 0.68 | 0.86 | 0.61 | 0.85 |
| | D-Tree | 0.52 | 0.78 | 0.63 | 0.82 |
| | SVM | 0.57 | 0.79 | 0.57 | 0.82 |

The results of applying rules in the document level are compared to the results of non-application of rules in the Table XV of the previous section, especially when $D_{Max\_Max}$ formula for sentence score computing and total 7 features are used. For example, the average F-score in positive classes for testing data in Table XV is **0.93.** While the average F-score in positive classes for testing data in case of applying rules is **0.94**, and the average F-score in negative classes for testing data in Table XV is **0.92.** While the average F-score in negative classes for testing data when rules are applied is **0.93**. From the given results, we conclude that the effectiveness of the rules appears clearly in the document level rather than in the term level of Arabic SA.

As a result, in the term level of Arabic SA, the best results are obtained by using $M_{Max\_Sub}$ formula to compute each term prior polarity score, then, applying total of 8 features with ANN classifier. While, in the document level, the best results are obtained by applying $D_{Max\_Max}$ formula for computing sentence score then, using total 7 features with ANN classifier. In other words, the best results are achieved by firstly, the application of rules for each term, then computing each sentence score using the $D_{Max\_Max}$ formula, and lastly, using a total of 7 features and an ANN classifier in the document level.

To evaluate the results of our study with respect to studies in literature, the best results obtained by Rushdi-Saleh et al. [13] are compared with our best results. It is important noting that OCA corpus and ML approach are used in both studies, and the best results are related to Precision, Recall and F-score evaluation measures. Although the Recall of Rushdi-Saleh' study (0.95) is better than the Recall of our study **(0.93)**, the Precision of our study (0.94) is better than the Precision of of Rushdi-Saleh' study (0.87), and that for F-score of our study **(0.94)** outperforms that of Rushdi-Saleh's study (0.91). In general, the results of this study are better than the results of Rushdi-Saleh's study.

## VI. Conclusion

We firstly aimed to enhance term level sentiment score by creating a new success formula to compute one prior polarity score for each sentiment term. Secondly, to improve the performance results of both term and document levels at Arabic SA by investigating the most commonly used ML classifier to classify the opinions embedded in the documents. Finally, implementation of the ML approach with rules is proposed to improve obtained SA performance values. According to experimental results, the usage $M_{Max\_Sub}$ formula is recommended for getting only one score for each Arabic sentiment term. Features have been extracted based on the term level and document level with application and non-application of rules in the term level. Document level SA is proposed to get better performance. The main findings achieved by our study is that, proposed feature vectors for document level SA gives the best performance with ANN classifier.